\documentclass[acmtog,nonacm]{acmart}
\acmSubmissionID{411}

\usepackage{booktabs} 
\usepackage[ruled]{algorithm2e} 

\SetAlFnt{\small}
\SetAlCapFnt{\small}
\SetAlCapNameFnt{\small}
\SetAlCapHSkip{0pt}

\usepackage{amsmath,amsthm}

\usepackage{amssymb}
\usepackage{graphicx}
\usepackage{textcomp}
\usepackage{wrapfig}
\usepackage{subfig}
\usepackage{color}
\usepackage{xspace}
\usepackage{overpic}
\usepackage{subfig}
\usepackage{enumitem}
\usepackage{booktabs}
\usepackage{hyperref}
\usepackage{comment}
\usepackage{multirow}
\usepackage{mathrsfs}

\newcommand{\neighborhood}{\mathscr{N}}

\acmJournal{TOG}
\acmVolume{43}
\acmNumber{4}
\acmArticle{83}
\acmYear{2024}
\acmMonth{7}

\setcopyright{acmlicensed}

\acmDOI{10.1145/3658169}

\begin{document}
\title{Spatial and Surface Correspondence Field for Interaction Transfer}

\author{Zeyu Huang}
\orcid{0000-0001-6786-1997}
\affiliation{%
 \institution{Shenzhen University}
 \country{China}}
\email{vcchzy@gmail.com}

\author{Honghao Xu}
\orcid{0009-0005-6380-4078}
\affiliation{%
 \institution{Shenzhen University}
 \country{China}}
\email{littledaisy20001227@gmail.com}

\author{Haibin Huang}
\orcid{0000-0002-7787-6428}
\affiliation{%
 \institution{Kuaishou Technology}
 \country{China}}
\email{jackiehuanghaibin@gmail.com}

\author{Chongyang Ma}
\orcid{0000-0002-8243-9513}
\affiliation{%
 \institution{Kuaishou Technology}
 \country{China}}
\email{chongyangm@gmail.com}

\author{Hui Huang}
\orcid{0000-0003-3212-0544}
\affiliation{%
 \institution{Shenzhen University}
 \country{China}}
\email{hhzhiyan@gmail.com}

\author{Ruizhen Hu}
\authornote{Corresponding author: Ruizhen Hu (ruizhen.hu@gmail.com).}
\orcid{0000-0002-6798-0336}
\affiliation{%
 \institution{Shenzhen University}
 \country{China}}
\email{ruizhen.hu@gmail.com}

\begin{abstract}
In this paper, we introduce a new method for the task of interaction transfer. Given an example interaction between a source object and an agent, our method can automatically infer both surface and spatial relationships for the agent and target objects within the same category, yielding more accurate and valid transfers. Specifically, our method characterizes the example interaction using a combined spatial and surface representation. We correspond the agent points and object points related to the representation to the target object space using a learned spatial and surface correspondence field, which represents objects as deformed and rotated signed distance fields. With the corresponded points, an optimization is performed under the constraints of our spatial and surface interaction representation and additional regularization. Experiments conducted on human-chair and hand-mug interaction transfer tasks show that our approach can handle larger geometry and topology variations between source and target shapes, significantly outperforming state-of-the-art methods. 
\end{abstract}

%
%


%
%

\begin{teaserfigure}
  \includegraphics[width=\textwidth]{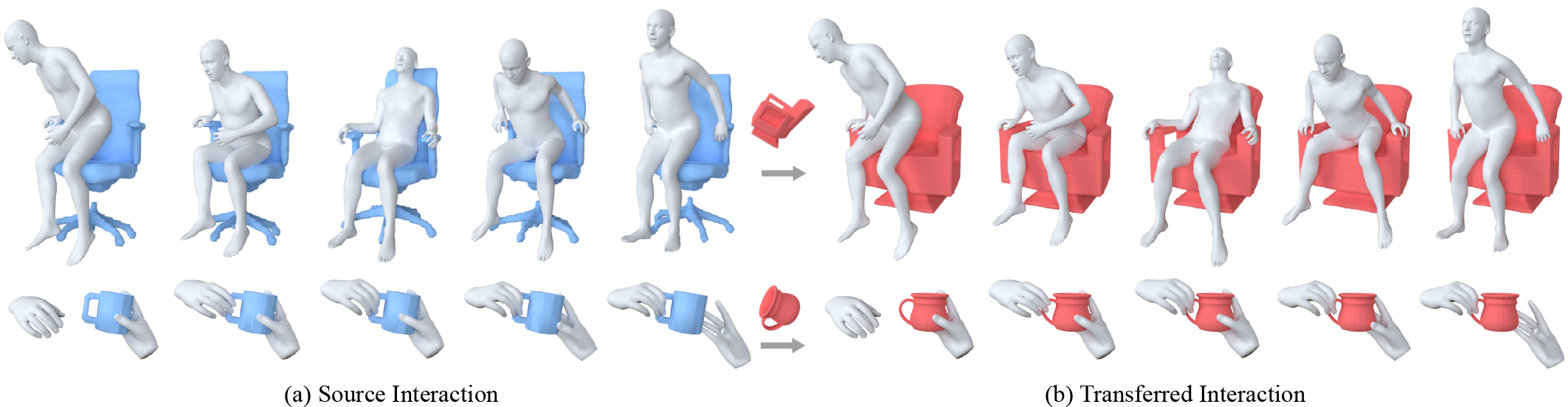}
  \caption{We present a novel method that leverages both spatial and surface correspondence to achieve accurate and valid interaction transfer results. Given the source interaction sequence shown on the left, our method can successfully transfer the agent (human body or hand) to perform similar interactions with the new object colored in red. }
  \label{fig:teaser}
\end{teaserfigure}

\maketitle

\section{Introduction}

Interactions play a fundamental role in understanding 3D objects and have wide-ranging applications in various fields such as games, movies, augmented reality/virtual reality, and robotics. However, enabling machines or agents to automatically infer their interactions with 3D objects and execute corresponding actions remains a challenging task. Traditional approaches involve manual interaction design by artists or recording interactions using motion capture devices, which are costly and not easily scalable.

In this work, we propose to tackle this problem with interaction transfer. Instead of inferring interactions directly, interaction transfer aims to empower an agent to predict and synthesize interactions by adapting existing interaction data to new objects that share similar geometric structures. By leveraging knowledge gained from observed interactions, agents can generalize and apply their understanding to previously unseen objects. This process is non-trivial and particularly challenging when dealing with deformable agents, such as parametric models of human bodies or hands, and transferring their interactions from a source object to a target object with a different shape but a similar pose.

Previous works have made progress in interaction transfer by leveraging dense surface correspondence to transfer contact points from the source object to the target object. The surface correspondence can be obtained by nearest neighbor searching, non-rigid registration~\cite{myronenko2010point}, explicit template matching~\cite{rodriguez2018transferring}, and shape interpolation in the latent space of neural implicit fields~\cite{yang2022oakink}. The transferred contact points are used to guide the optimization of the agent pose. However, contact points are usually local and sparse, making these methods sensitive to the surface correspondence results.

To address this limitation, \citeN{kim2016retargeting} propose to build a spatial correspondence to directly map all agent points to the desired position, which requires mesh inputs and complicated pre-processing. In contrast, \citeN{simeonov2022neural} propose to optimize the global pose of the agent directly in the learned neural descriptor field of respective objects by feature matching without explicit point correspondence. These methods provide global and continuous point transfer but do not explicitly consider the local details.

To overcome all these challenges, we utilize both spatial and surface interaction representations to encode global and local features of interactions. We develop a unified surface and spatial correspondence method by leveraging the neural implicit field. Specifically, given a source interaction and a target object, our method maps them into a learned template field. In the field, the interacted regions on the source object's surface are corresponded to the target object's surface, and the spatial coordinates of the agent points are corresponded from the source space. The agent parameters are then optimized with the interaction representation as well as the correspondence constraints, enabling the transfer of interactions from the source object to the target object.

To validate the effectiveness of our proposed method, we conduct experiments on various interaction transfer tasks, including human-chair and hand-mug interaction transfers. Our approach demonstrates superior performance compared to state-of-the-art methods based on either contact points or neural descriptor fields. The ablation studies further demonstrate the importance of jointly modeling the interaction with spatial and surface representations. These findings validate the effectiveness and practicality of our proposed approach for accurate and realistic interaction transfers.

To summarize, our contributions are as follows:
\begin{itemize}
\item We introduce spatial and surface interaction representation (SSIR), a novel approach to encapsulate both global and local features of interactions.
\item We propose spatial and surface correspondence field (SSCF), a unified framework for establishing both surface and spatial correspondences between two shapes.
\item We develop spatial and surface constrained optimization (SSCO), a method ensuring accurate and valid interaction transfers.
\end{itemize}

\section{Related Work}

Interaction transfer and pose transfer have been extensively studied in the field of computer graphics. Numerous works have focused on transferring the pose from one character to another character with a different skeleton, shape, or scale, while the characters interact with the same object. This problem is commonly referred as motion retargeting~\cite{ho2010spatial,al2013relationship,kim2021interactive, jin2018aura,liu2018surface,basset2019contact,aberman2019learning,aberman2020skeleton,zhang2023simulation}. In contrast, the interaction transfer discussed in our work assumes that the agent remains the same, but the interacting object changes. The objective is to transfer the agent's pose from one object to another in order to achieve a similar spatial relationship between the agent and the object. 

\paragraph{Interaction transfer.}
Existing approaches for interaction transfer often require an interaction representation and an optimization based on that representation. One commonly used representation is contact points on the object surface that are close to the agent during the interaction~\cite{rodriguez2018transferring,yang2022oakink,wu2023functional}. The goal of contact points optimization is to keep these points in contact or maintain the same distances relative to the agent as the source interaction. However, these approaches are sensitive to the surface correspondence result. To overcome this limitation, \citeN{kim2016retargeting} proposed representing the agent interaction as spatial points in the space of the object, and establishing a spatial map to correspond the source spatial points to desired locations in the space of the target object. While spatial points provide accurate localization, complex pre-processing is required to convert the source and target objects to genus-zero watertight meshes, which poses challenges for large-scale interaction transfer and cannot handle target objects with partial observations. More recently, \citeN{simeonov2022neural} introduced the neural descriptor field (NDF) for interaction transfer. NDF represents the interaction as a set of spatial points and their associated features in an SE(3)-equivariant neural descriptor field. Interaction transfer is achieved through feature matching between the NDFs of the source and target objects. However, NDF is learned based on point occupancy relative to the object and designed to perform interaction transfer for simple two-finger grippers. This limits its ability to provide detailed shape correspondence for complex interaction transfer involving deformable agents such as human bodies or hands.
\begin{figure*}[!t]
    \centering
    \includegraphics[width=\textwidth]{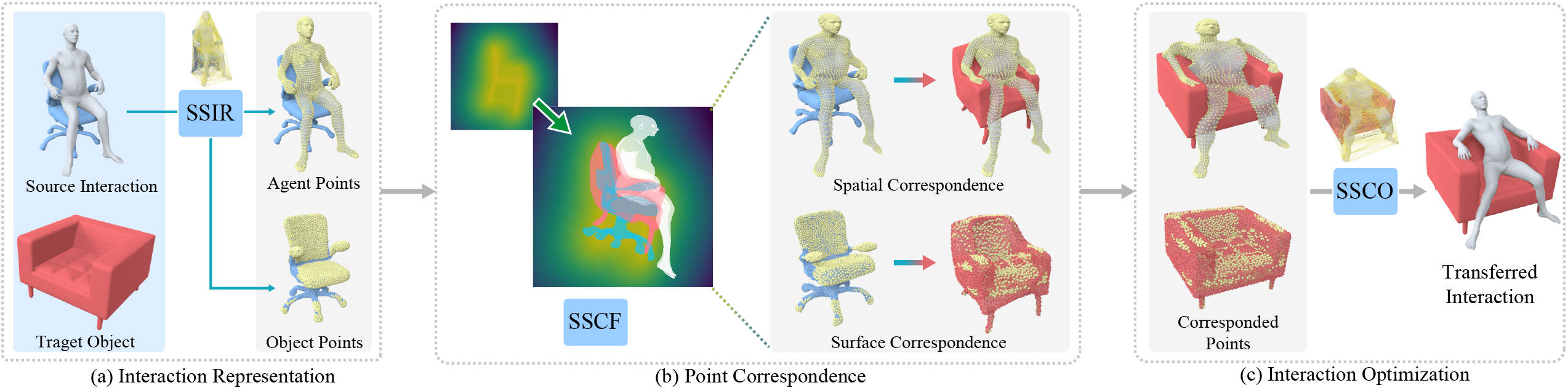}
    \caption{Method overview. (a) Given a source interaction, our method characterizes it with a combination of spatial and surface representations using the agent points and part of the object points. (b) To transfer the interaction to a target object, we map the agent points and object points related to our interaction representation to the target object space using a learned spatial and surface correspondence field. (c) With the corresponded agent points and object points, we perform an optimization constrained by the spatial and surface representation of the interaction in order to transfer the agent to the target object.}
    \label{fig:overview}
\end{figure*}

\paragraph{Geometric representation of interaction.}
Apart from interaction representations specially designed for interaction transfer tasks, there have also been other geometric interaction representations that provide informative interaction information. \citeN{ho2010spatial} represent the entire interaction scene as an interaction mesh and preserve the topological relationship between agents using Laplacian coordinates in motion retargeting.
\citeN{zhang2023simulation} further use the edges and edge distances of the interaction mesh to learn a control policy for multi-agent interactions in physical environments. \citeN{al2013relationship} represent agent joints as a linear combination of a set of environment points for agent motion retargeting when the environment deforms. \citeN{zhao2014IBS} introduce the interaction bisector surface (IBS) to measure the topological and geometric similarity of interactions for retrieval. IBS is also utilized for interaction completion and generation~\cite{zhao2017character, zhao2019localization}.
\citeN{huang2023nift} propose combining IBS with a more sophisticated neural field than NDF~\cite{simeonov2022neural} in order to capture the spatial relationship between the agent and the object and guide object manipulation. However, this method only works for rigid agents. \citeN{pirk2017understanding} develop a histogram representation to describe the spatial and temporal characteristics of agent trajectories, which is also only utilized for interaction classification and object retrieval. 

\paragraph{Shape correspondence.}
Since our method requires building both spatial and surface correspondence between two shapes provided in arbitrary poses, we also discuss works on shape correspondence. Early methods use non-rigid registration~\cite{li2008global,myronenko2010point} to establish dense correspondences between two point clouds. These methods require an initial coarse alignment of the two objects to avoid local minima. Functional maps propose computing correspondences between functions defined on the shapes to find dense surface correspondence between two objects~\cite{ovsjanikov2012functional,ren2021discrete,magnet2022smooth}. Template methods~\cite{zheng2021deep,deng2021deformed,sun2022topology,kim2023semantic} learn a deformation model from a template shape for all shapes in the same category, so that the surface correspondences are naturally created. We extend the representation of ~\cite{deng2021deformed} to arbitrarily rotating objects and focus more on its ability in spatial representation rather than the surface correspondence.  

\section{Method}

As shown in Figure~\ref{fig:overview}, our method characterizes an interaction with a combination of spatial and surface representations using agent points and part of the object points. To transfer such a source interaction to a target object, we learn a spatial and surface correspondence field to correspond the spatial points and surface points in the source object space to the target object space with the power of the neural implicit field. Finally, an optimization of the agent is performed with the corresponded points constrained by the interaction representation to obtain the transferred interaction on the target object.

\paragraph{Notations.} We denote an object point cloud as $O=\{o_i \mid i=0,...,n_O\}$ and an deformable agent as $A(\Pi,\Theta)=\{a_i \mid i=0,...,n_A\}$ parameterized by rigid transformation $\Pi$ and joint angles $\Theta=\{\theta_i \mid i=0,...,n_J\}$. Given a source interaction $(O^s,A^s)$ and a target object $O^t$, our goal is to find the optimized agent parameters $A^t = (\Pi^t,\Theta^t)$ such that the interaction between $O^t$ and $A^t$ is similar to the source interaction between $O^s$ and $A^s$.

\subsection{Spatial and Surface Interaction Representation}
To describe the source interaction $(O^s, A^s)$ and facilitate interaction transfer, we represent the interaction with a combination of spatial and surface representation, named SSIR. 

To encode the global position of the interacting agent in the space of the interacted object, we use the spatial coordinates of the agent points relative to the object center $A^s-c^s$ as the spatial part of SSIR. We assume all objects are centered at the origin, so the spatial representation can be simply denoted as $A^s$. 

To further capture the local geometric details of each point $a^s_i$ on the agent surface, we use the graph based Laplacian coordinate $\delta^s_i$ as the surface part of SSIR, which is demonstrated in~\cite{zhou2005large} to preserve the local topology and avoid self-interaction during deformation. More specifically, we perform Delaunay tetrahedralization on all points in $(O^s,A^s)$ and use the edges of the tetrahedrons to build the graph connections $G$.
The Laplacian coordinate $\delta^s_i$ is defined as the local difference between each agent point $a^i$ and a linear combination of its neighboring points in the graph $G$ :
\begin{gather}
\delta^s_i = \mathscr{L}_G(a^s_i)=a^s_i-\sum_{j \in \neighborhood_A(i)} w_{ij}a^{s}_j-\sum_{j \in \neighborhood_{O^s}(i)} w_{ij}o^{s}_j \\
\text{s.t.} \sum_{j \in \neighborhood_A(i)} w_{ij}+\sum_{j \in \neighborhood_{O^s}(i)} w_{ij} = 1 \notag
\end{gather}
where $\neighborhood_A(i)$ and $\neighborhood_{O^s}(i)$ are the graph neighborhoods of $a^s_i$ in the agent and the source object points respectively, and $w_{ij}$ is a normalized weight in inverse proportion to the edge lengths. 

To transfer the source interaction to a target object $O^t$,  we need to correspond the spatial coordinates of the agent from the space of the source object to the target object, and their graph neighborhoods on the surface of the source object to the target object, which then can be used to guide the interaction optimization.

\begin{figure*}[!t]
    \centering
    \includegraphics[width=\textwidth]{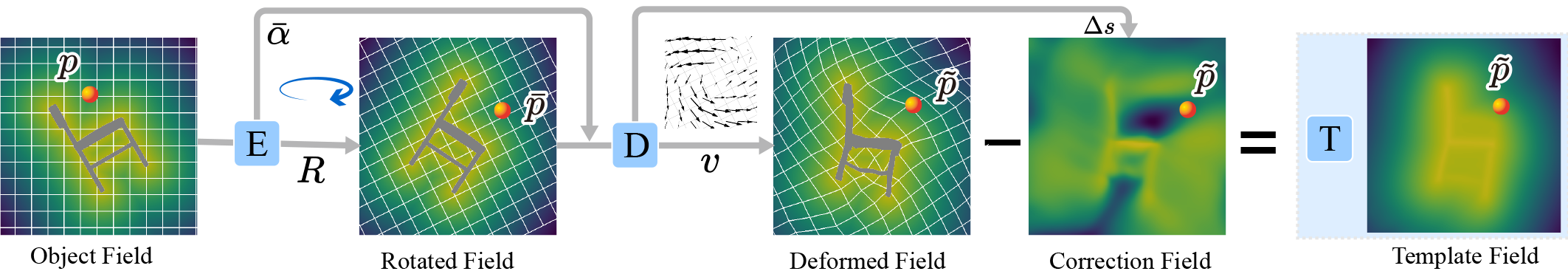}
    \caption{Object representation. Our method represents the object signed distance field (SDF) as a rotated and deformed neural implicit field. Given an object input, our encoder $E$ encodes it into a rotation-invariant shape code $\bar \alpha$ and predicts a rotation matrix $R$, which rotates points $p$ in the object field to canonical posed points $\bar p$. Conditioned on the shape code, our decoder $D$ corresponds $\bar p$ to the template field $\tilde{p}$ by estimating the deformation $v$ and outputs an SDF correction $\Delta s$. The SDF of $p$ is defined as the SDF of $\tilde p$ in the template field $T(\tilde p)$ plus the correction value $\Delta s$.}
    \label{fig:net}
\end{figure*}

\subsection{Spatial and Surface Correspondence Field}
\label{sec:learning}
To build such a spatial and surface correspondence between the source object and the target object, we learn an implicit object representation, as illustrated in Figure~\ref{fig:net}. 

Our shape representation is based on the deformed implicit field (DIF)~\cite{deng2021deformed}. DIF represents the signed distance field (SDF) of a 3D object with two neural fields $T$ and $D$:
\begin{gather}
    T: p \rightarrow s ,\\
    D: (p,\alpha) \rightarrow (v,\Delta s) ,
\end{gather}
where $T$ is a shared template SDF field for objects of the same category, and $D$ is a deformation and correction field conditioned on a shape code $\alpha$. Given a query point $p$ in the space of the object represented by $\alpha$, $D$ predicts a point-wise deformation $v$ and SDF correction $\Delta s$. The SDF of $p$ defined by DIF is $s=T(p+v)+\Delta s$. 

\paragraph{RDIF representation} We further propose to learn a rotated and deformed implicit field (RDIF) that is also valid for randomly rotated objects. Given an object point cloud $O$ as input, our RDIF adds an additional shape encoder $E$:
\begin{equation}
    E: O  \rightarrow (\bar \alpha,R) ,
\end{equation}
where $\bar \alpha$ is a rotation-invariant shape code of $O$, and $R$ is a rotation matrix to align the randomly rotated $O$ with the template field $T$. The SDF of a query point $p$ defined by RDIF is $s=T(Rp+v)+\Delta s$. 

In order to generate the same rotation-invariant shape code $\bar \alpha$ for any randomly rotated instances of the same object $O$, the shape encoder $E$ consists of two submodules $E_\alpha$ and $E_R$:
\begin{gather}
    E_\alpha: O \rightarrow  \alpha, \\
    E_R: \alpha  \rightarrow R,
\end{gather}
where both $E_\alpha$ and $E_R$ are implemented with rotation-equivariant Vector Neuron (VN) layers~\cite{deng2021vector}. The input and output of VN layers are all matrices of 3D vectors, allowing the rotation matrix applied to the input to be transmitted across the network by $E_\alpha(OR^T)=E_\alpha(O)R^T$. Then the network $E_R$ is used to extract the rotation matrix $R$ from the output of $E_\alpha$, and the rotation-invariant shape code is constructed by $\bar \alpha = E_\alpha(OR^T)R  = E_\alpha(O)$.  

The network of RDIF is trained in a self-supervised manner so that the rotation matrix does not require ground truth for supervision. More specifically, for any rotated query point $Rp$ of $p$ in the space of the rotated object $OR^T$ of $O$, they share the same ground truth SDF. The supervision of SDF can encourage the network to learn a canonical pose for the same object or all objects of the same category, and predict the correct rotation matrix for each input to align with the canonical pose. We use the same training losses as DIF~\cite{deng2021deformed} to train our RDIF. Refer to the supplementary material for more details about the network architecture and training.

Note that the design of RDIF is tailored for our interaction transfer task. 
Compared to DIF, RDIF adds a rotation-invariant shape encoder to enable dealing with shapes given in arbitrary poses that are quite common in the interaction transfer task. Moreover, we further explore and utilize the spatial correspondence learned with the template field to provide more accurate and spatial control of the interaction transfer, while DIF only uses the deformation field as a surface correspondence field to register object surfaces.

\paragraph{Point correspondence} We use the learned template field of RDIF as our spatial and surface correspondence field, or SSCF, to correspond the source agent points $A^s$ and object points $O^s$ to the space and surface of the target object $O^t$ as $(A^{st},O^{st})$. Given a query point $p$ in the input object field, we denote the rotated and deformed point in the template field as $\tilde p=\bar p+v=Rp+v$, as shown in Figure~\ref{fig:net}. 

For object points, we correspond a point $o^s_i$ on the surface of the source object to a point $o^{st}_i$ on the surface of the target object with the minimum distance in SSCF:
\begin{equation}
o^{st}_i = o^{t}_j, \text{ where } j = \arg \min_{j} ||\tilde{o}^s_i - \tilde{o}^t_j||.
\end{equation}

Likewise, for agent points, we correspond a point $a^s_i$ in the space of the source object to a point $a^{st}_i$ in the space of target object. Other than selecting the corresponded point from target object points, a set of densely sampled spatial points $X^t$ around the target object is used to compute the distance in SSCF:
\begin{equation}
a^{st}_i = x^{t}_j, \text{ where } j = \arg \min_{j} ||\tilde{a}^s_i - \tilde{x}^t_j||.
\end{equation}

\subsection{Spatial and Surface Constrained Optimization}
\label{sec:optimization}
Given the source interaction $(A^s,O^s)$ and the corresponded points $(A^{st},O^{st})$, we introduce our spatial and surface constrained optimization on the parameter space $\{\Theta,\Pi\}$ of the agent to transfer the interaction to the target object, named SSCO. We use the following terms to guide SSCO:

\paragraph{Spatial loss.} We use the corresponded agent points as direct position references for the current agent points $A(\Theta,\Pi)$:
\begin{equation}
    L_{\text{spatial}}(\Theta,\Pi) = \sum_{i}||a_i-a^{st}_i||^2.
\end{equation}

\paragraph{Surface loss.} We use the source Laplacian coordinates to preserve the local details of the current agent surface relative to $O^{st}$:
\begin{equation}
    L_{\text{surface}}(\Theta,\Pi) = \sum_{i}||\mathscr{L}_G(a_i)-R_i\delta^{s}_i||^2,
\end{equation}
where $R_i$ is a rotation matrix to align the normal of $a^s_i$ to $a_i$ since the Laplacian coordinates are not rotation invariant. 

\paragraph{Penetration loss.} We detect object points inside the current agent as $O^{t_{in}}$ and minimize their distances to their nearest agent points:
\begin{equation}
    L_{\text{pen}}(\Theta,\Pi) = \sum_{o \in O^{t_{in}}} \min_i||a_i-o||^2.
\end{equation}

\paragraph{Smooth term.} For a sequence of interaction, we add a smooth term to consider temporal continuity between the agent poses next to each other:
\begin{equation}
    L_{\text{smooth}}(\Theta_k,\Theta_{k-1})=||\Theta_k-\Theta_{k-1}||^2.
\end{equation}

Finally, the overall transfer loss for a single interaction is:
\begin{equation}
    L_{\text{SSCO}}(\Theta,\Pi) = \lambda_1 L_{\text{spatial}}+\lambda_2 L_{\text{surface}}+ \lambda_3 L_{\text{pen}},
\end{equation}
and the optimization is formulated as:
\begin{gather}
    \Theta^{t},\Pi^{t}  =  \arg \min_{\Theta,\Pi} L_{\text{SSCO}}(\Theta,\Pi) \\
    \text{s.t.} \quad |\theta_i-\theta^s_i|  \le  \gamma_{\theta}, \notag 
\end{gather}
where $\lambda_1$, $\lambda_2$, and $\lambda_3$ are loss weights, and $\gamma_\theta$ is a threshold to limit the change of each joint angle relative to the source pose. We find setting loss weights to $1$ and $\gamma_\theta=10^\circ$ is sufficient for all experiments.

For interaction sequences, we optimize a batch of $K$ interactions at the same time:
\begin{equation}
    \Theta^{t}_{1,...,K},\Pi^{t}_{1,...,K}  =  \underset{\Theta_{1,...,K},\Pi_{1,...,K}}{\arg\min} \sum_{k} L_{\text{SSCO}}(\Theta_k,\Pi_k) + \lambda_4L_{\text{smooth}}, 
\end{equation}
where $\lambda_4=0.01$ is the weight of the smooth term. We optimize a batch of $K=12$ interactions at the same time and use a sliding window with a stride of six frames for the entire sequence.

\section{Experiments}
In this section, we present qualitative results and comparisons to demonstrate the effectiveness of our method. We showcase visual examples of interaction transfer accomplished by our approach, highlighting the improved quality and accuracy compared to existing methods. Furthermore, we analyze the results and evaluate the contributions of various components within our approach.

\subsection{Experimental Settings}

\paragraph{Dataset.} In our experiments, we primarily focus on the transfer of human-chair and hand-mug interactions. For human-chair interactions, we utilize interactions from the CHAIRS dataset~\cite{jiang2022chairs} as the source interactions, which use SMPL-X~\cite{SMPL-X:2019} as the parametric model of human agents. For hand-object interactions, we use interactions from the OakInk-Core dataset~\cite{yang2022oakink} as the source interactions, which uses MANO~\cite{MANO:SIGGRAPHASIA:2017} as the parametric model of hands. For the target objects, we randomly select objects of the same category from the ShapeNet dataset~\cite{shapenet2015} and chair scans from the ScanNet dataset~\cite{dai2017scannet}. These target objects provide a diverse set of shapes and functionalities for evaluating the effectiveness of our method in interaction transfer tasks.

\paragraph{Implementation details.}
To train our network, we randomly split $90\%$ of the chairs and mugs from the ShapeNet dataset as the training set, and the remaining $10\%$ are used as the test set. Data augmentation techniques are applied, including random scaling in the range of $[0.75, 1.25]$ and random rotations in $SO(3)$. The network is trained using Adam optimizer~\cite{kingma2014adam} with an initial learning rate of $0.0001$, which is halved every 100 epochs. For spatial point correspondence, we use grid points with an edge resolution of 128 in the space of the target object as $X^t$. For interaction optimization, we use gradient descent with a learning rate $0.01$, which is halved every 10 iterations. Gradient clipping is performed with a maximum gradient norm of $0.01$. The optimization is stopped when the total iteration reaches 100 or the overall loss changes by less than $0.0001$. 

\begin{figure}[!t]
    \centering
    \includegraphics[width=\columnwidth]{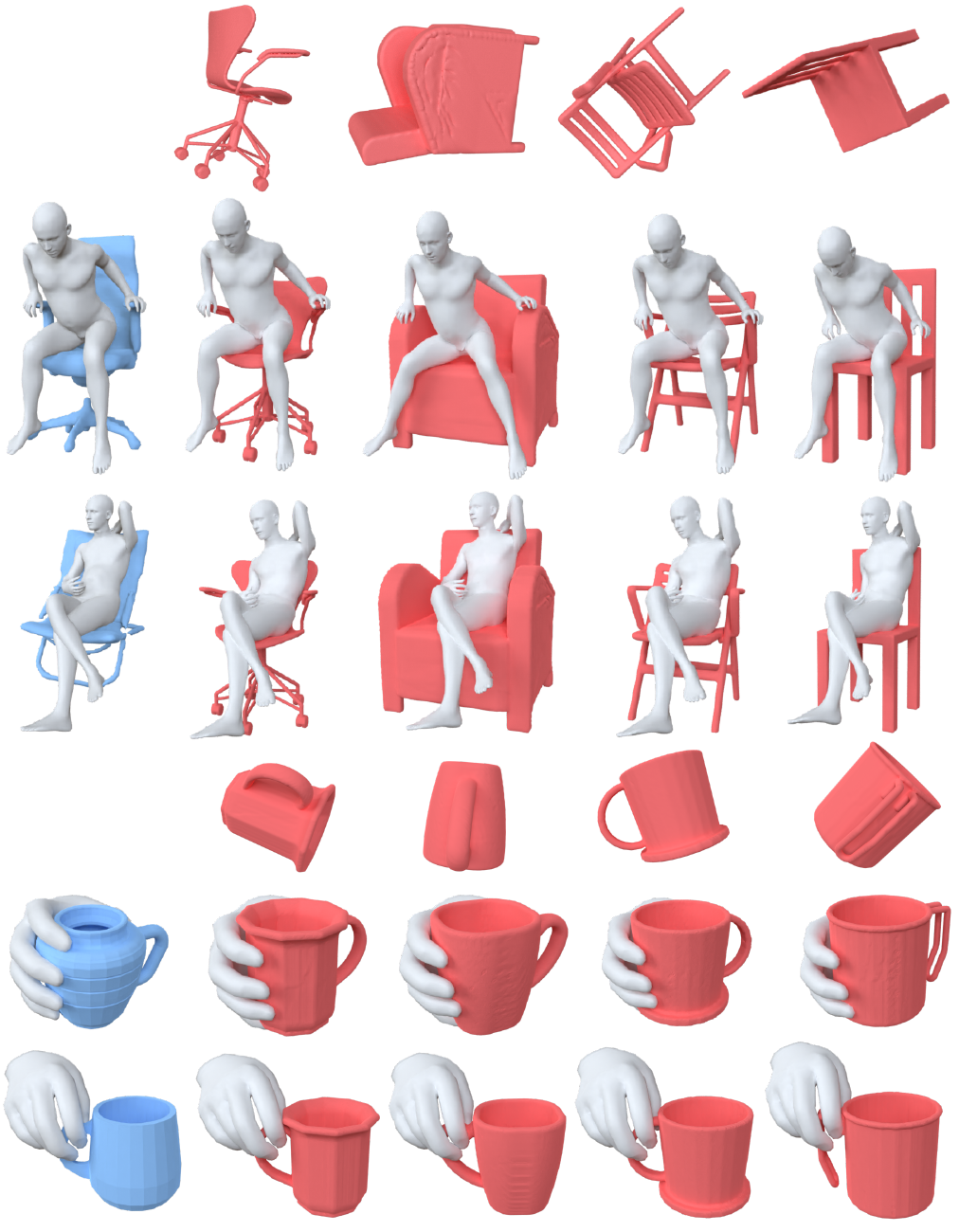}
    \caption{Static interaction transfer results. Our method allows the transfer of interactions to objects with various structures, geometries, and arbitrary orientations.}
    \label{fig:seq}
\end{figure}

\begin{figure*}[!t]
    \centering
    \includegraphics[width=0.96\textwidth]{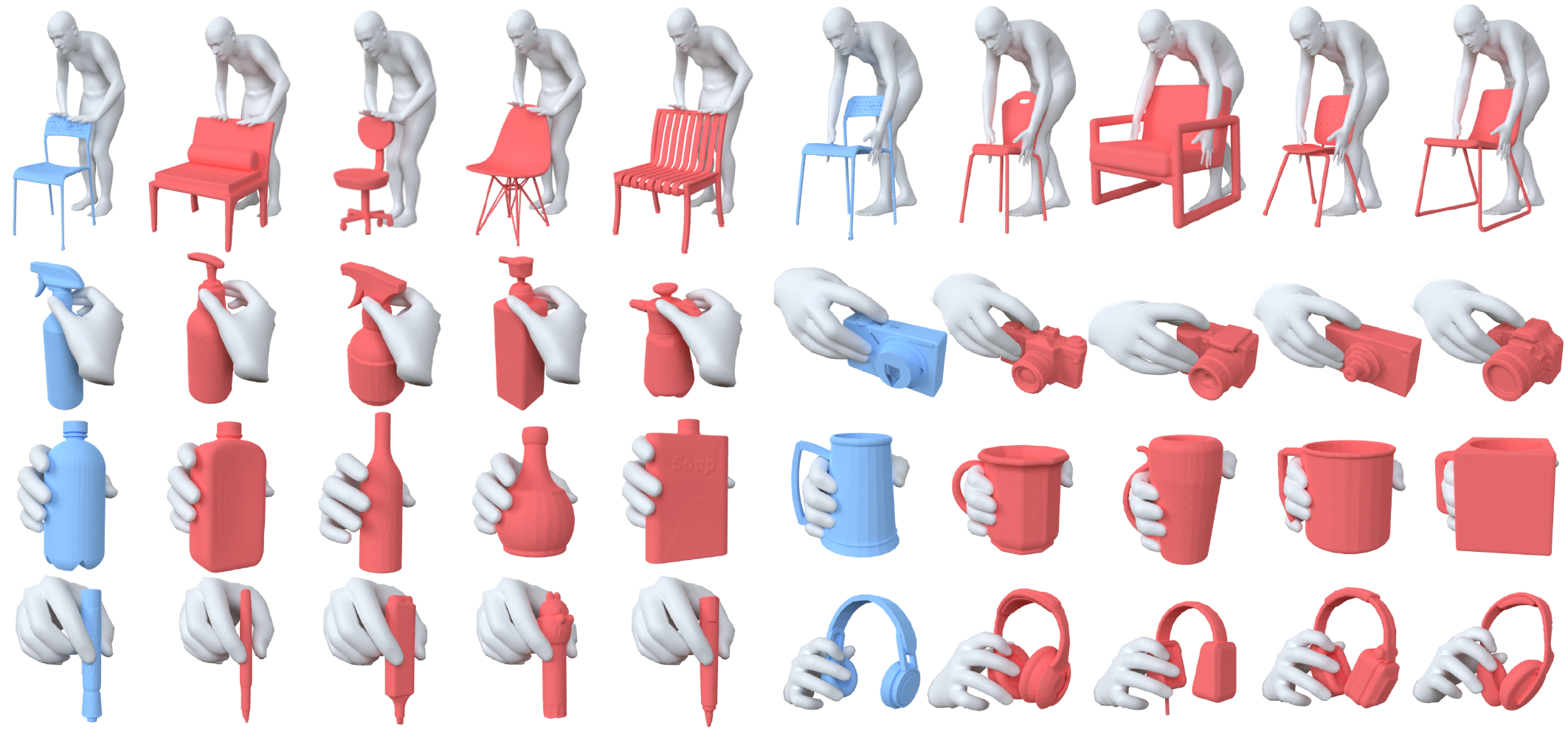}
    \caption{More results of our method on static interaction transfer.}
    \label{fig:more_static}
\end{figure*}

\subsection{Qualitative Evaluation}
We demonstrate the results of interaction transfer under various scenarios, including static and motion sequence transfer for synthetic scenes as well as interactions with real scans. 

\paragraph{Static interaction transfer.}

Figure~\ref{fig:seq} (Top) showcases human-chair interactions transferred using our method. The left side of the figure shows the source interactions, while the top side presents the target objects. Notably, our method handles objects with arbitrary orientations, correctly aligning all target objects with the source object using our learned canonical template implicit field. We observe that our method successfully transfers various interaction poses across chairs with different structures and geometries. For example, our method accurately transfers the contact between the hand and armrest 
to other chairs. Additionally, our method effectively transfers the interaction between the buttocks and seat to chairs with different geometry and topology. 
Similarly, Figure~\ref{fig:seq} (Bottom) showcases several examples of hand-mug interactions transferred using our method. Notably, our approach accurately preserves the important interaction semantics when transferring to mugs with different shapes and structures.
We provide additional examples in Figure~\ref{fig:more_static} to demonstrate that our method supports more types of human-chair interactions from the OMOMO dataset~\cite{li2023object} and more categories of hand-object interactions from the OakInk dataset~\cite{yang2022oakink}.

\paragraph{Interaction sequence transfer.}

Figure~\ref{fig:teaser} shows the results of transferring interaction sequences from the source object to the target objects. Notably, our method presents adaptive changes in hand contacts based on the positions of armrests in different chairs during human-chair interactions. This demonstrates the ability of our method to accurately transfer interactions while considering the specific characteristics of the target objects. Furthermore, our method successfully preserves the consistency of two agents in the two-hand interaction transfer with the mug, ensuring both hands interact with the mug in a consistent and visually plausible manner. Refer to supplementary material for motion results.

\begin{figure}[!t]
    \centering
    \includegraphics[width=0.95\columnwidth]{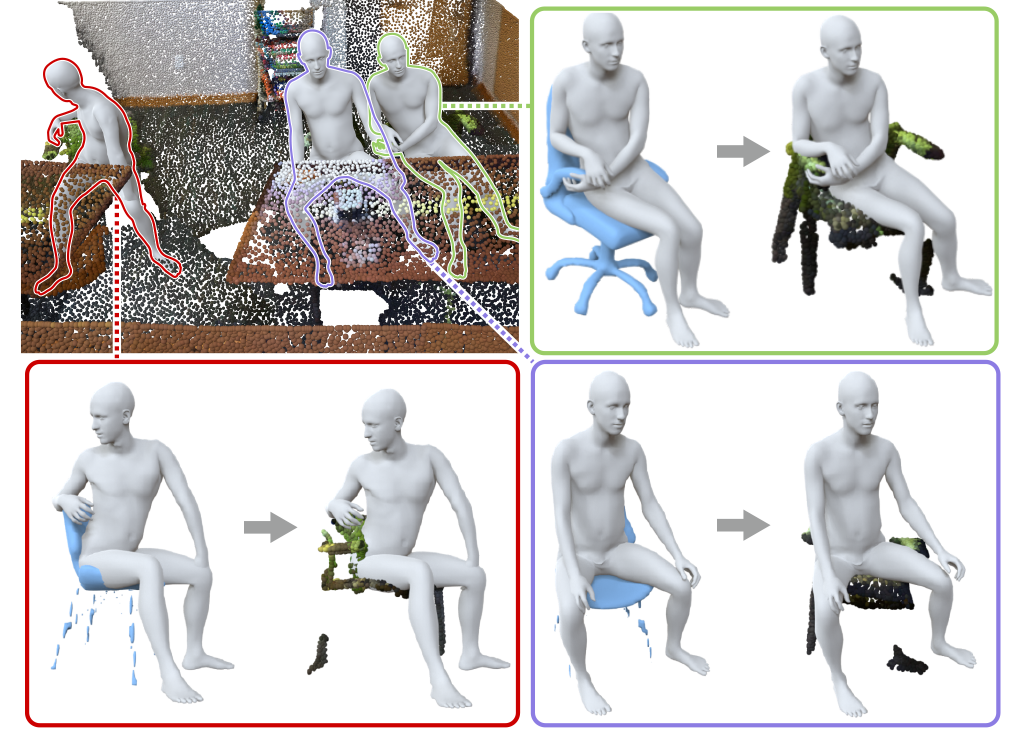}
    \caption{Interaction transfer with real scan inputs. Our method is capable of transferring interactions to scanned objects, enabling varied and realistic interaction transfer in real-world scenarios. }
    \label{fig:scan}
\end{figure}

\begin{figure*}[!t]
    \centering
    \includegraphics[width=0.97\textwidth]{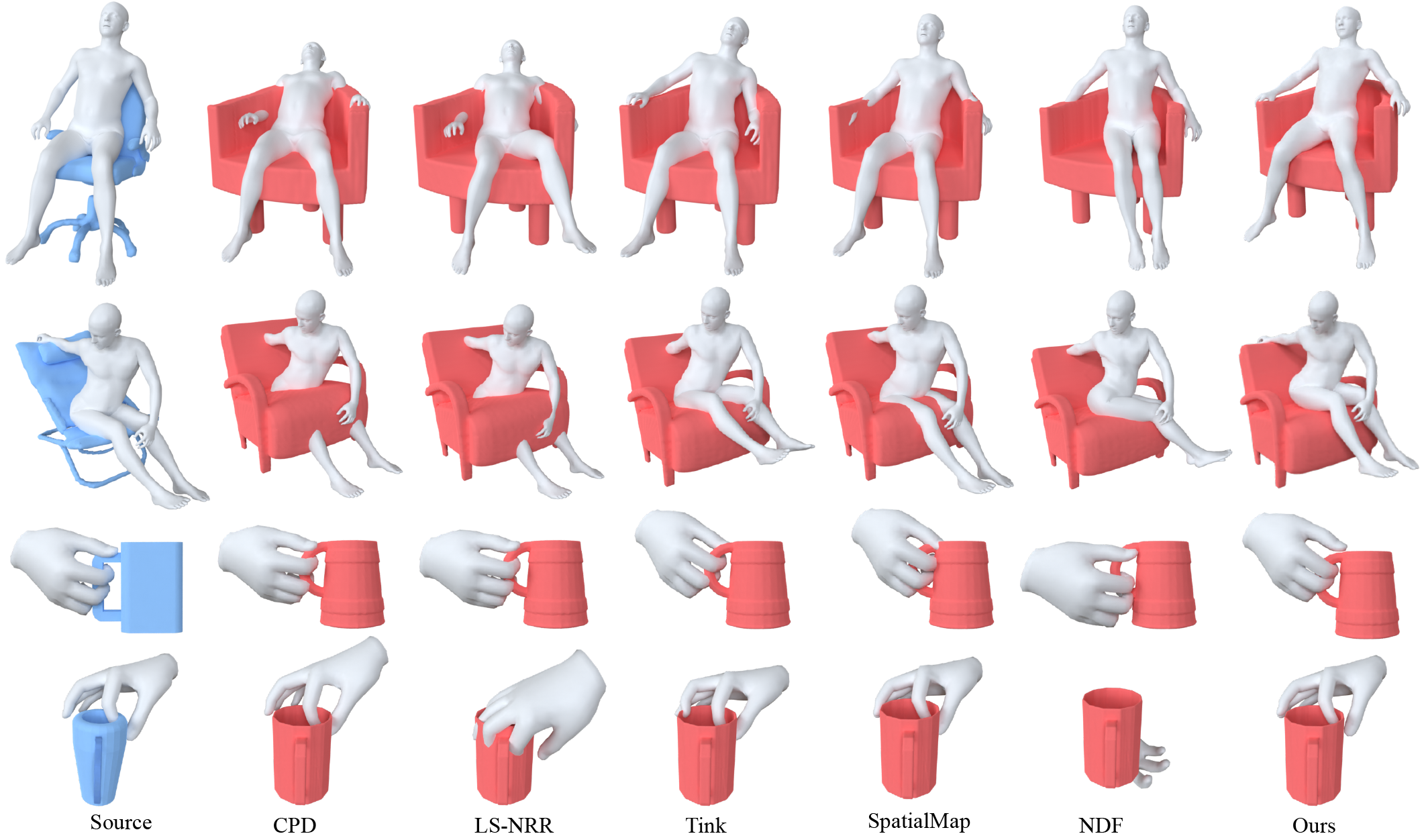}
    \caption{We compare our method with baseline methods in both human-chair and hand-mug interaction transfer. Our method presents superior results in accurately transferring interactions between objects in terms of visual quality and consistency. }
    \label{fig:comp_chair}
\end{figure*}

\paragraph{Real scan transfer.}
We present the results of transferring interactions to real scan scenes in Figure~\ref{fig:scan}. For each segment of chairs, couches, and sofas in the ScanNet dataset, we randomly select a source interaction. Without retraining on real scan data, our network effectively aligns the real chair segments with the source objects and successfully transfers the interactions into the scene. The results demonstrate the versatility and adaptability of our method in handling real-world scan data.  Furthermore, we observe the diversity in which the agents interact with the scene, illustrating the ability of our method to generate various and realistic interaction behaviors.
Please refer to our supplemental materials for motion results.

\begin{table}
\centering
\caption{Quantitative comparison on human-chair interaction transfer.}
\label{tab:chair}
\resizebox{\columnwidth}{!}{
\begin{tabular}{c|l|c|c|c|c} 
\hline
\multicolumn{2}{c|}{Method} & Dep. $\downarrow$ & Vol. $\downarrow$ & IoU $\uparrow$ & Time $\downarrow$ \\ 
\hline
\multirow{4}{*}{\begin{tabular}{c}Surface \\ based\end{tabular}} 
 & CPD~\cite{myronenko2010point} & 0.81 & 8.01 & 15.6 & 30.1 \\
 & LS-NRR~\cite{rodriguez2018transferring} & 0.85 & 9.22 & 13.8 & 8.2 \\
 & Tink~\cite{yang2022oakink} & 0.84 & 6.98 & 15.9 & 25.6 \\
 & Tink (ours) & 0.83 & 6.30 & 15.5 & \textbf{2.9} \\ 
\hline
\multirow{2}{*}{\begin{tabular}{c}Spatial \\ based\end{tabular}} 
 & SpatialMap~\cite{kim2016retargeting} & 0.80 & 6.49 & 15.5 & 9.7 \\
 & NDF~\cite{simeonov2022neural} & 0.69 & 4.23 & 15.8 & 8.0 \\ 
\hline
\multicolumn{2}{c|}{Ours} & \textbf{0.49} & \textbf{3.18} & \textbf{21.3} & \underline{4.3} \\
\hline
\end{tabular}}
\end{table}

\subsection{Comparison Results}
\paragraph{Baselines.} We conduct a comprehensive comparison with traditional and state-of-the-art methods: CPD~\cite{myronenko2010point}, LS-NRR~\cite{rodriguez2018transferring}, Tink~\cite{yang2022oakink}, SpatialMap~\cite{kim2016retargeting}, and NDF~\cite{simeonov2022neural}. Additionally, we propose a new baseline by replacing the surface correspondence used in Tink with ours and refer it as ``Tink (ours)''. It is important to note that, except for NDF, all the other methods assume that the source object and target object are initially aligned in the canonical pose. For a fair comparison, we provide the object rotation learned by our method to these methods. 

\begin{table}
\centering
\caption{Quantitative comparison on hand-mug interaction transfer.}
\label{tab:mug}
\resizebox{\columnwidth}{!}{
\begin{tabular}{c|l|c|c|c|c} 
\hline
\multicolumn{2}{c|}{Method} & Dep. $\downarrow$ & Vol. $\downarrow$ & IoU $\uparrow$ & Time $\downarrow$ \\ 
\hline
\multirow{4}{*}{\begin{tabular}{c}Surface \\ based\end{tabular}} 
& CPD~\cite{myronenko2010point} & 0.80 & 5.16  & 13.4 & 28.5 \\
& LS-NRR~\cite{rodriguez2018transferring} & 0.78 & 5.19  & 13.2 & 3.8 \\
& Tink~\cite{yang2022oakink}  & 0.75 & 5.18  & 12.2 & 25.8 \\ 
& Tink (ours)  & 0.72 & 4.93  & 11.9 & \textbf{1.4} \\ 
\hline
\multirow{2}{*}{\begin{tabular}{c}Spatial \\ based\end{tabular}} 
& SpatialMap~\cite{kim2016retargeting} & 0.68 & 4.48  & 12.4 & 7.5 \\
& NDF~\cite{simeonov2022neural} & 0.63 & 3.19  & 7.9 & 4.0 \\ 
\hline
\multicolumn{2}{c|}{Ours} & \textbf{0.56} & \textbf{2.99} & \textbf{14.1} & \underline{3.2}\\
\hline
\end{tabular}}
\end{table}

\begin{figure*}[!t]
    \centering
    \includegraphics[width=0.96\textwidth]{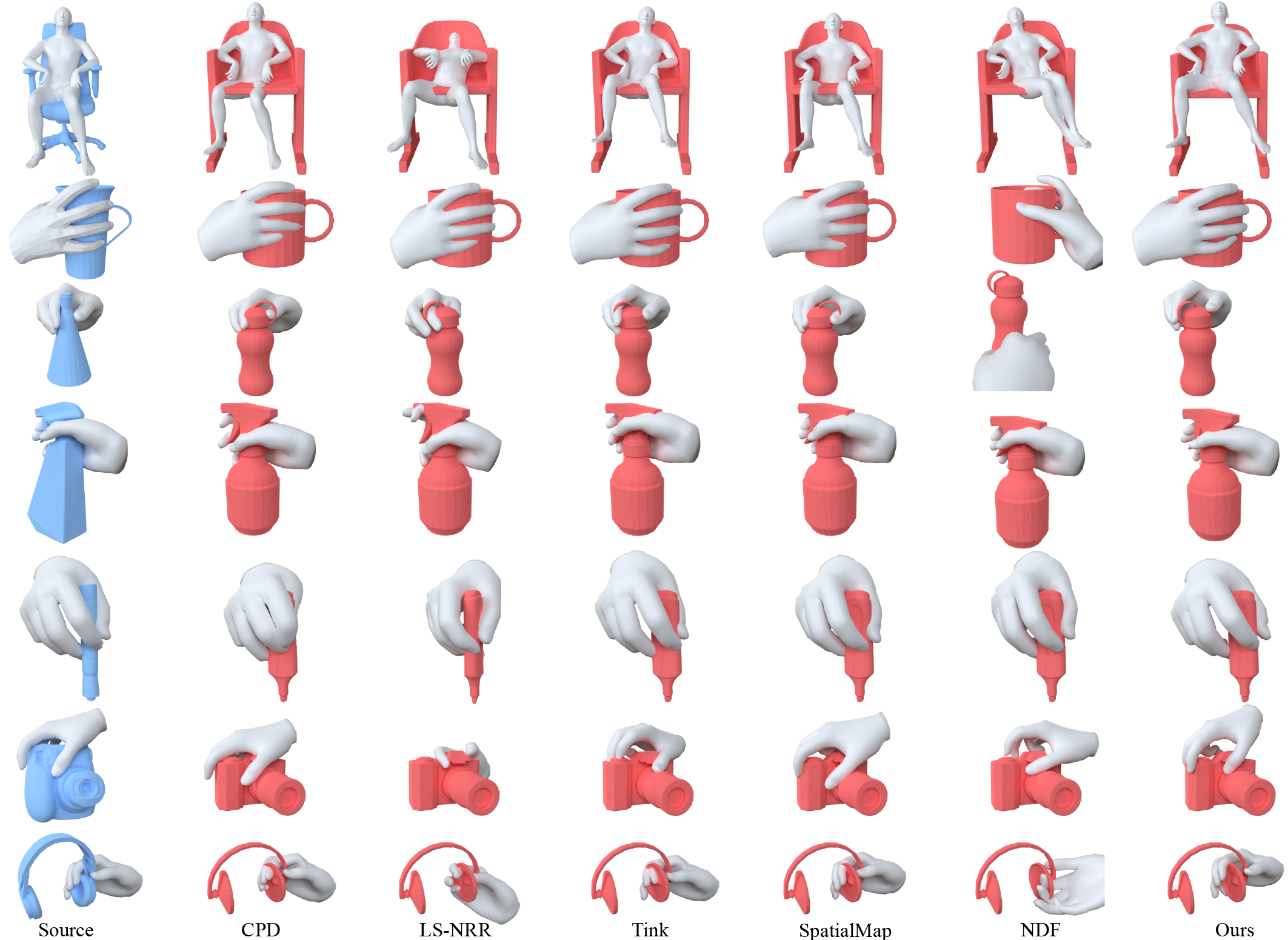}
    \caption{More comparisons of our method against baseline methods on various interactions.}
    \label{fig:more_comp}
\end{figure*}

\begin{table*}
\centering
\caption{Quantitative comparison on other OakInk interactions.}
\label{tab:oakink}
\resizebox{\linewidth}{!}{
\begin{tabular}{c|l|ccc|ccc|ccc|ccc|ccc|ccc}
\hline
\multicolumn{2}{c|}{\multirow{2}{*}{Method}} & \multicolumn{3}{c|}{Cylinder Bottle} & \multicolumn{3}{c|}{Trigger Sprayer} & \multicolumn{3}{c|}{Pen} & \multicolumn{3}{c|}{Camera} & \multicolumn{3}{c|}{Headphone} & \multicolumn{3}{c}{Average} \\
\cline{3-20}
\multicolumn{2}{l|}{} & Dep.~$\downarrow$ & Vol.~$\downarrow$ & IoU~$\uparrow$ & Dep.~$\downarrow$ & Vol.~$\downarrow$ & IoU~$\uparrow$ & Dep.~$\downarrow$ & Vol.~$\downarrow$ & IoU~$\uparrow$ & Dep.~$\downarrow$ & Vol.~$\downarrow$ & IoU~$\uparrow$ & Dep.~$\downarrow$ & Vol.~$\downarrow$ & IoU~$\uparrow$ & Dep.~$\downarrow$ & Vol.~$\downarrow$ & IoU~$\uparrow$ \\ 
\hline
\multirow{4}{*}{\begin{tabular}[c]{@{}c@{}}Surface \\ based\end{tabular}} & CPD~\cite{myronenko2010point} & 0.47 & 0.65 & 8.87 & 0.83 & 2.33 & 6.54 & 0.93 & 0.13 & 15.30 & 0.84 & 1.90 & 11.90 & 0.57 & 0.26 & 11.70 & 0.73 & 1.05 & 10.86 \\
 & LS-NRR~\cite{rodriguez2018transferring} & 0.47 & 0.58 & 10.20 & 0.98 & 3.19 & 4.57 & 0.61 & 0.04 & 22.20 & 0.94 & 2.25 & 6.63 & 0.65 & 0.33 & 10.70 & 0.73 & 1.28 & 10.86 \\
 & Tink~\cite{yang2022oakink} & 0.36 & 0.33 & 9.76 & 0.77 & 1.54 & 6.31 & 0.53 & 0.04 & 25.30 & \underline{0.70} & 1.11 & 12.80 & 0.54 & \underline{0.23} & 11.80 & 0.58 & 0.65 & 13.19 \\
 & Tink (ours) & 0.35 & 0.35 & 9.88 & 0.79 & 1.61 & 6.28 & 0.54 & 0.04 & 25.00 & 0.74 & 1.03 & 10.50 & 0.57 & 0.24 & 11.50 & 0.60 & 0.65 & 12.63 \\ 
\hline
\multirow{2}{*}{\begin{tabular}[c]{@{}c@{}}Spatial \\ based\end{tabular}} & SpatialMap~\cite{kim2016retargeting} & 0.37 & 0.31 & \textbf{14.30} & 0.78 & 1.13 & \underline{7.00} & 0.53 & 0.03 & \textbf{36.70} & 0.71 & 1.11 & \underline{13.90} & \underline{0.48} & 0.28 & \underline{13.70} & 0.57 & 0.57 & \underline{17.12} \\
 & NDF~\cite{simeonov2022neural} & 0.28 & \textbf{0.20} & 12.00 & \underline{0.63} & \underline{0.93} & 6.38 & \textbf{0.33} & \textbf{0.01} & 33.60 & 0.71 & \underline{0.64} & 6.75 & 0.57 & 0.45 & 8.99 & \underline{0.50} & \underline{0.45} & 13.54 \\ 
\hline
\multicolumn{2}{c|}{Ours} & \textbf{0.25} & \underline{0.21} & \underline{12.50} & \textbf{0.51} & \textbf{0.70} & \textbf{9.05} & \underline{0.41} & \underline{0.02} & \underline{34.50} & \textbf{0.48} & \textbf{0.43} & \textbf{15.90} & \textbf{0.46} & \textbf{0.16} & \textbf{14.00} & \textbf{0.42} & \textbf{0.30} & \textbf{17.19} \\
\hline
\end{tabular}}
\end{table*}

\paragraph{Evaluation metrics.} We use evaluation metrics in OakInk~\cite{yang2022oakink} to assess the geometric correctness of the transferred interactions, including penetration depth and solid intersection volume. To evaluate the similarity of the interactions, we compute the Intersection over Union (IoU) of contact between the source and the target agents. To determine the contact points, we consider agent points that are within a certain distance threshold from the object. For the hand-mug interaction, the distance threshold is set to $2mm$, while for the human-chair interaction, we set it to $2cm$. We also compute the average transfer time in seconds for each method.

\paragraph{Protocol.} We randomly sample 10 interactions from CHAIRS~\cite{jiang2022chairs} and OakInk~\cite{yang2022oakink} as the source interactions. We randomly sample 100 mugs or chairs from ShapeNet as target objects.
For other categories, since only a few objects are available in the OakInk dataset~\cite{yang2022oakink}, the maximum target number for each category is set to 50.
Target chairs are sampled from the test set not seen by the network and other categories are sampled from the entire set due to the limited number of objects available. All quantitative comparisons are run on a machine with an Intel E5 CPU, a 12 GB NVIDIA TITAN V GPU, and 128 GB RAM.

\begin{figure*}[!t]
    \centering
    \includegraphics[width=0.96\textwidth]{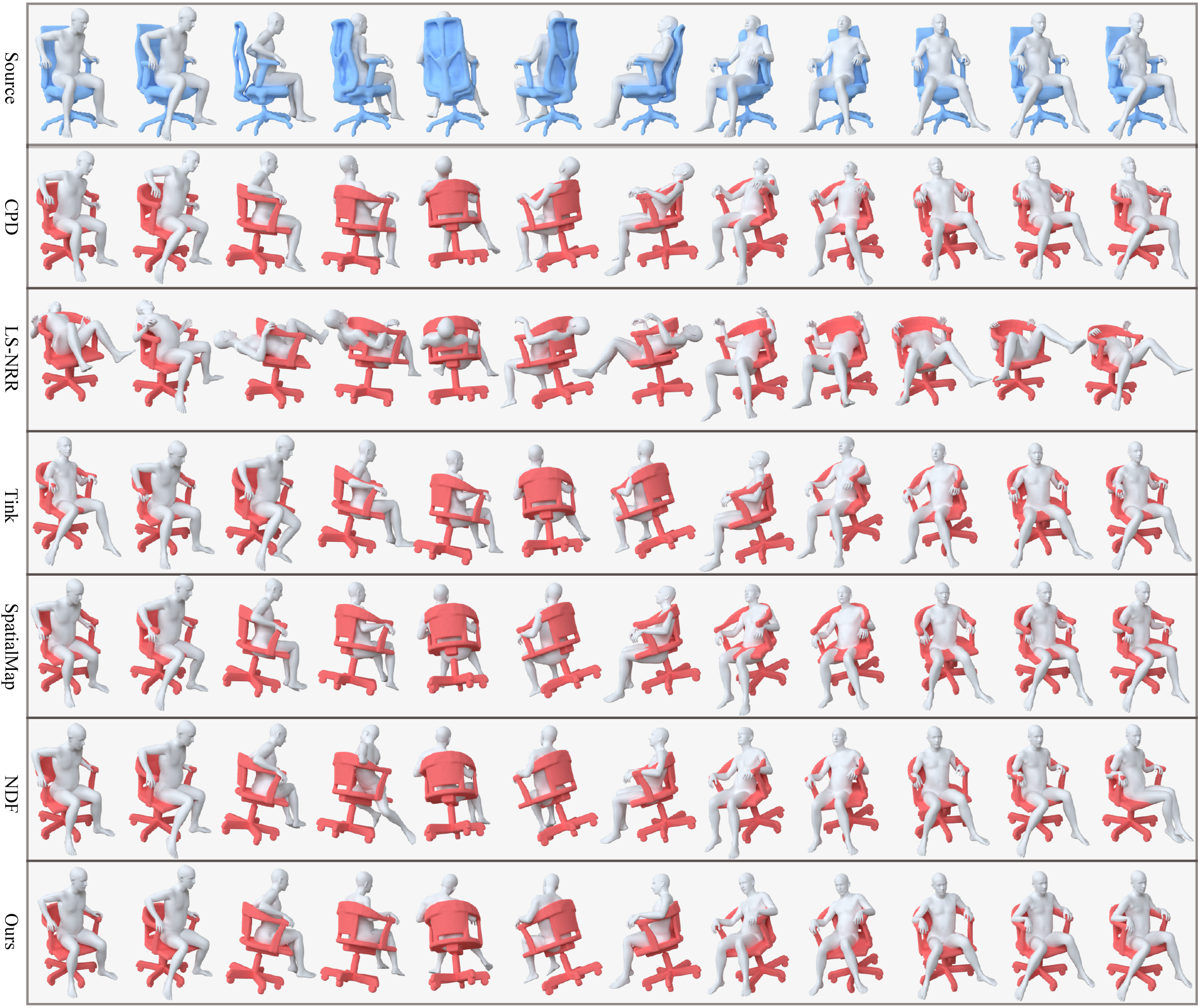}
    \caption{Qualitative comparison on sequence transfer. The source agent sits and rotates the chair.}
    \label{fig:comp_motion}
\end{figure*}

\paragraph{Quantitative comparison.}
Tables~\ref{tab:chair} and \ref{tab:mug} present the quantitative comparisons of our method with all the baseline methods. When evaluating the penetration depth and intersection volume, we observe that methods solely relying on contact points correspondence on the object surface, such as CPD, LS-NRR, and Tink, perform worse than methods that incorporate spatial correspondence, such as SpatialMap, NDF, and our approach. Furthermore, we can eliminate the impact of different surface correspondence methods by comparing Tink and Tink (Ours). Among the methods utilizing spatial correspondence, our method demonstrates superior performance with a significant margin in penetration metrics while also maintaining the highest contact IoU. This indicates that our transferred interactions are not only more accurate in terms of object penetration but also exhibit a higher similarity to the source interactions. Due to efficient spatial and surface correspondence, our method and our proposed Tink variant also have the lowest time costs. Quantitative results for other categories of interactions from the OakInk dataset~\cite{yang2022oakink} are shown in Table~\ref{tab:oakink}. SpatialMap~\cite{kim2016retargeting} and NDF~\cite{simeonov2022neural} perform well on interactions of simple objects, such as cylinder bottle and pen, but struggle on interactions of complex objects, such as trigger sprayer, camera, and headphone. However, our method provides results with the most balanced penetration metrics and contact IoU for all cases and achieves the best average performance.

\paragraph{Qualitative comparison.} 

\begin{figure*}[!t]
    \centering
    \includegraphics[width=0.95\textwidth]{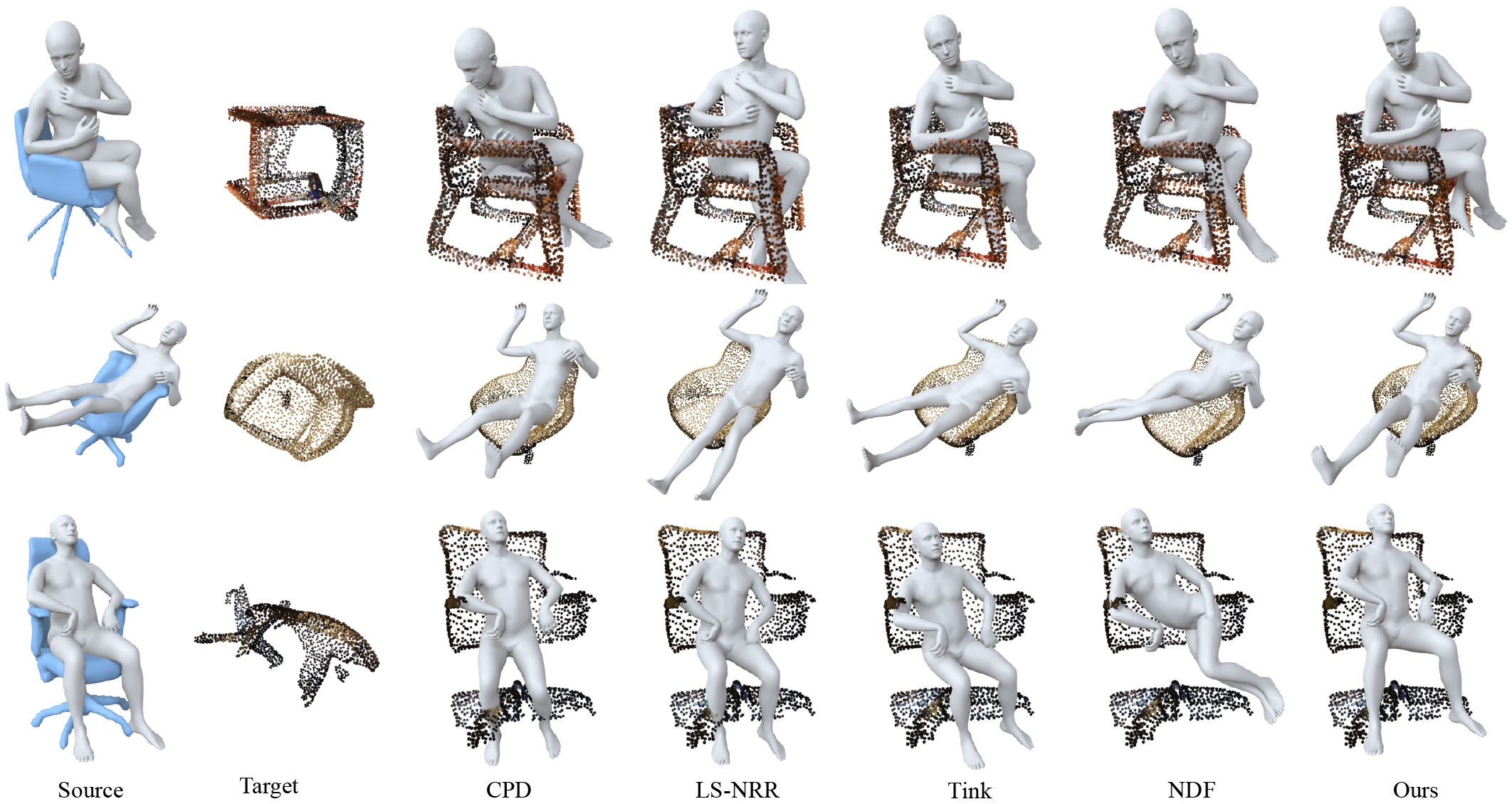}
    \caption{Qualitative comparison on real scan inputs.}
    \label{fig:comp_scan}
\end{figure*}

Figures~\ref{fig:comp_chair},~\ref{fig:comp_motion}, and~\ref{fig:comp_scan} provide qualitative comparisons between our method and the baselines on static interaction, interaction sequence, and real scan inputs. When comparing our method to the state-of-the-art neural descriptor field method (NDF), we observe that our transferred interactions exhibit a more consistent agent orientation with the source interaction. Furthermore, when comparing our method to the baseline method that uses spatial correspondence (SpatialMap), we find that our results exhibit fewer instances of object penetration. Additionally, SpatialMap requires mesh inputs which prevents its application to partial scans. Compared to the baselines that solely rely on surface correspondence, our method produces transferred interactions with body part placements that are more similar to the source interaction. Finally, our method can also produce more stable and smooth motions for motion sequences and more robust results on real scan inputs. Qualitative results for more categories of interactions from the OakInk dataset~\cite{yang2022oakink} are shown in Figure~\ref{fig:more_comp}. Notice the interaction details in our results, indicating that our method can generate more natural transferred interactions compared to other methods.

\subsection{Ablation Studies}

We further conduct ablation studies of our method on human-chair interaction to validate each component of our method and the results are presented in Table~\ref{tab:ablation}. 
When we replace the rotation-invariant encoder $E$ with a vanilla point cloud encoder like PointNet~\cite{qi2017pointnet}, the network becomes a simple DIF~\cite{deng2021deformed} with an encoder outputting a shape code and a rotation matrix. We see that both the penetration metrics and contact IoU deteriorate. It means DIF itself cannot effectively learn a valid template field from training data with arbitrary rotations, which is crucial for application in real scans. The inclusion of $L_{\text{pen}}$ increases the intersection volume but slightly improves the contact similarity. To intuitively see the effect of the other two losses in the optimization, we conduct experiments with only $L_{\text{spatial}}$ and only $L_{\text{surface}}$ respectively. 

\begin{table}
\caption{Ablation studies of our method on human-chair interaction.}
\label{tab:ablation}
\centering
\begin{tabular}{l|cc|c|c}
\hline
Ablation & Dep. $\downarrow$ & Vol. $\downarrow$ & IoU $\uparrow$ & Time $\downarrow$\\ 
\hline
Vanilla $E$ & 0.79 & 5.06 & 12.9 & 4.2 \\ 
Without $L_{\text{sdf}}$  & 0.52 & 4.33 & 21.9 & 4.1\\
$L_{\text{spatial}}$ only & 0.60 & 5.63 & \textbf{25.7} & \textbf{2.6} \\
$L_{\text{surface}}$ only & 0.51 & 3.44 & 19.2 & 3.3 \\
\hline
Ours & \textbf{0.49} & \textbf{3.18} & 21.3 &  4.3 \\
\hline
\end{tabular}
\end{table}

\begin{figure}
    \centering
    \includegraphics[width=\linewidth]{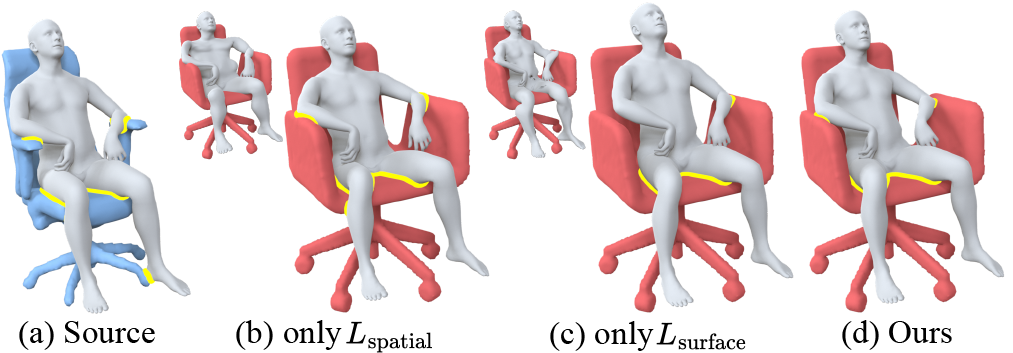}
    \caption{Ablation studies of losses. (a) Source interaction. (b) Corresponded agent points (left) and optimized agent pose (right). (c) Deformed agent points (left) and optimized agent pose (right). (d) Result of our full method. Mesh intersect edges are highlighted in yellow.}
    \label{fig:comp_loss}
\end{figure}

As shown in Figure~\ref{fig:comp_loss}(b), $L_{\text{spatial}}$ provides direct position guidance to preserve the spatial relationship between the object and agent, achieving high contact similarity with a relatively poor penetration depth and volume. 
Our method with only $L_{\text{surface}}$ can be considered as an extension of ~\cite{ho2010spatial} from motion targeting to interaction transfer. It uses corresponded object points to drive the agent deformation and tries to keep the topology of the interaction with graph-based Laplacian coordinates to avoid local intersection. 
In Figure~\ref{fig:comp_loss}(c), we show the optimized agent points before (left) and after (right) considering the kinematic model of the agent. We can see that the deformation caused by $L_{\text{surface}}$ is sparse and local, resulting in a lower penetration agent pose but less similarity in contact compared to the source interaction. As shown in Figure~\ref{fig:comp_loss}(d), our method achieves the best balance between contact similarity and geometric rationality.

\section{Conclusion}

In this work, we present a novel method for interaction transfer that enables an agent to predict and synthesize interactions with unseen objects. Our method can handle large topology changes of objects, since we use a neural implicit field to build the spatial and surface correspondence between the source and target objects.

\begin{figure}
    \centering
    \includegraphics[width=\linewidth]{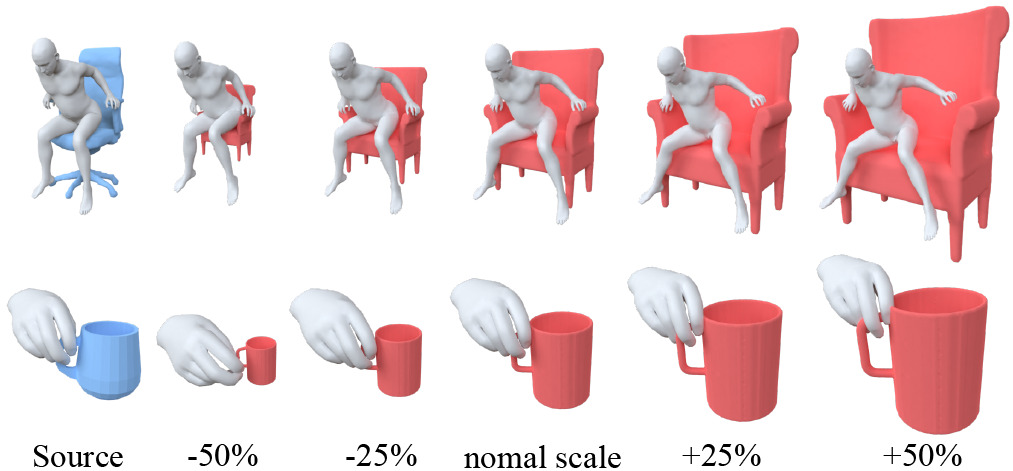}
    \caption{Interaction transfer results between objects of different scales.}
    \label{fig:scale}
\end{figure}

\begin{figure}
	\centering
	\includegraphics[width=\linewidth]{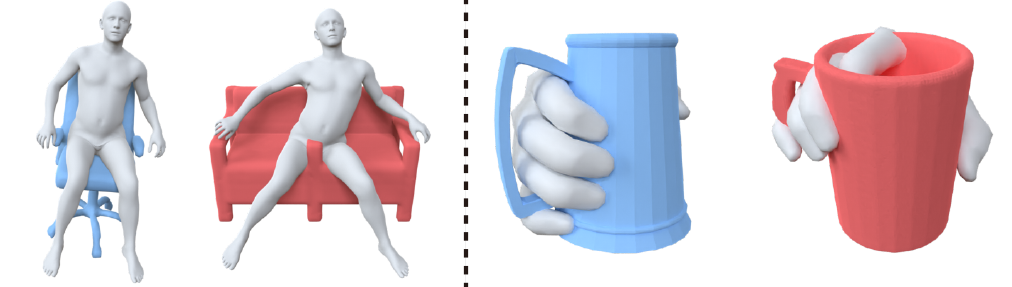}
	\caption{Some failure cases of our method.}
        \label{fig:failure_cases}
\end{figure}

\paragraph{Limitations and future work.}
We assume that interaction transfer is performed among objects of similar scales. Our method handles scale changes between -25\% and 25\% for smaller or larger target objects quite well, as shown in Figure~\ref{fig:scale}, since our network is trained with scale augmentation within this range. 
For scale changes beyond this range, our method may fail to generate plausible interactions in order to accommodate the uncommon shape, as in the case of the chair, or may fail to provide accurate correspondence for unseen scales, as in the case of the mug.
In addition, our method may face difficulties with objects that have the same topology but significantly different geometries in their corresponding interaction regions.
For example, as shown in Figure~\ref{fig:failure_cases}, our method generates suboptimal solutions for transferring the sitting pose from a one-seat chair to a two-seat sofa whose middle armrest is improperly corresponded to the seat.
Similarly, it struggles with the grasping pose transfer from a mug to another mug with a much smaller handle, which does not provide enough space to establish correspondence for three fingers.
A more advanced semantic or functional understanding of the source interaction may help alleviate this issue. It is worth exploring solutions to address these challenges and improve the robustness of our approach in the future.

\begin{acks}
We thank the anonymous reviewers for their valuable comments. This work was supported in parts by NSFC (62322207, 62161146005, U2001206), Guangdong Natural Science Foundation (2021B1515020085), Shenzhen Science and Technology Program (RCYX20210609103121030), DEGP Innovation Team (2022KCXTD025),  Guangdong Laboratory of Artificial Intelligence and Digital Economy (SZ) and Scientific Development Funds of Shenzhen University.
\end{acks}

\bibliographystyle{ACM-Reference-Format}
\bibliography{reference}

\appendix
\section{Network details}

\begin{figure}[!t]
    \centering
    \includegraphics[width=\linewidth]{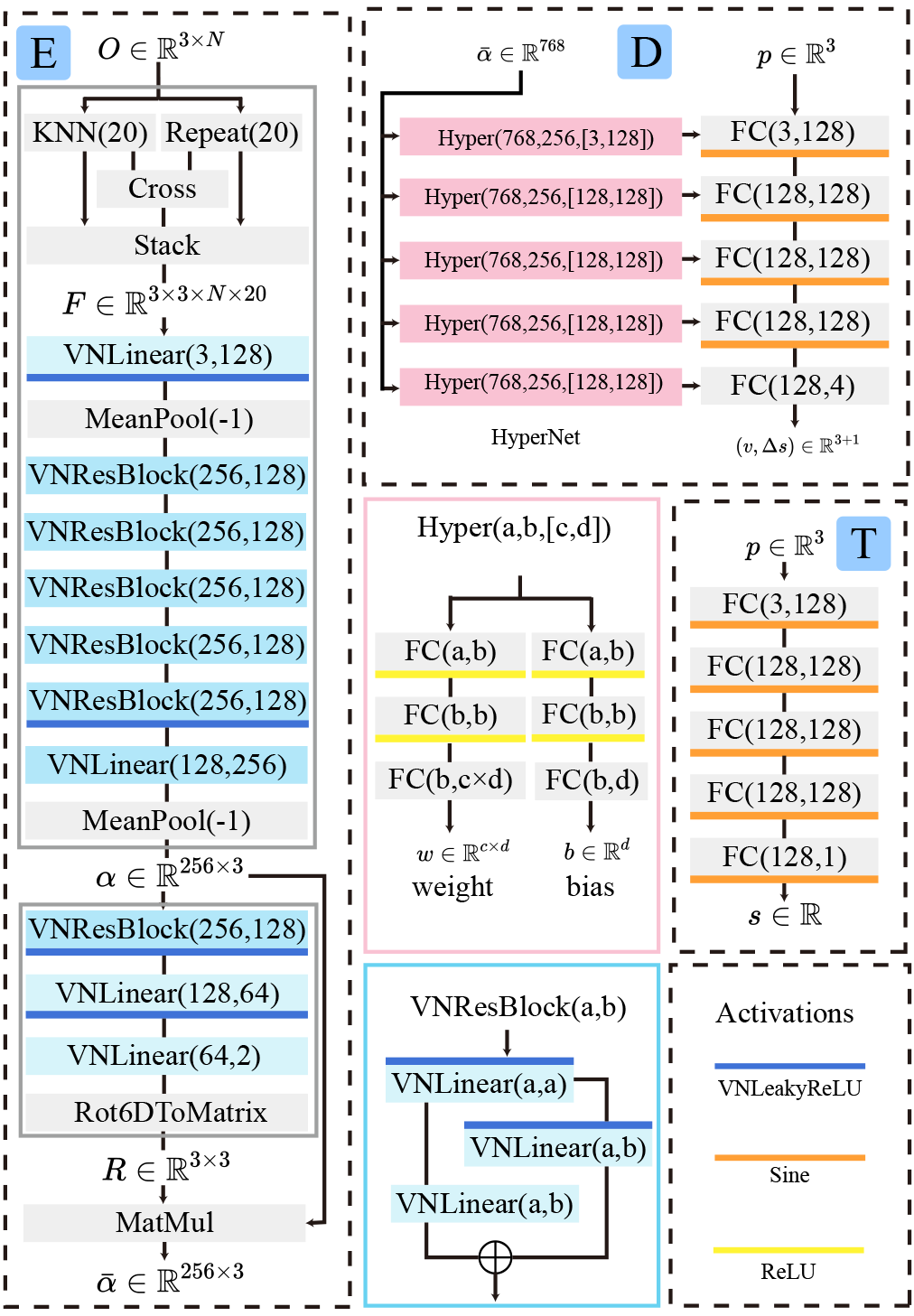}
    \caption{Network architecture.}
    \label{fig:archi}
\end{figure}

\paragraph{Network architecture}
Our network contains a shape encoder $E$, a deformation and correction decoder $D$, and a template decoder $T$. The architecture of each module is shown in Figure~\ref{fig:archi}. We implemented with the encoder $E$ with VNLinear layer and VNLeakyReLU layer in Vector Neuron layers~\cite{deng2021vector}, which can keep the input and output tensor rotation-equivariant. The deformation and correction decoder $D$ is a 5-layer MLP whose weights and bias are predicted with another network called HyperNet. The template decoder $T$ is also a 5-layer MLP with learnable weights. 

\paragraph{Training losses} Given an object $O$ and a query point $p$ with ground truth SDF $\hat s$, our network outputs a shape code $\bar \alpha$, a rotation matrix $R$, a deformation offset $v$, a SDF correction value $\Delta s$ and a template SDF $s$, we use the same loss as~\cite{deng2021deformed} to train our network:
\begin{equation}
    L_{net} = L_{sdf}+w_1L_{normal}+w_2L_{smooth}+w_3L_{c}+w_4L_{reg},
\end{equation}
where $w_{1,...,4}$ are weights for different losses. We set $w_1=1e2$, $w_2=5$, $w_3=1e2$, and $w_4=1e6$ for all object categories.

The SDF loss is defined as:
\begin{align}
        L_{sdf} & = \sum_i (\sum_{p \in \Omega}|\Phi_i(p)-\hat s|  + \sum_{p \in \mathcal{S}_i}(1-\langle \nabla \Phi_i(p),n \rangle) \\
        &+\sum_{p \in \Omega} | \| \nabla\Phi_i(p) \|_2-1|+\sum_{p \in \Omega \setminus  \mathcal{S}_i}exp(-\delta \cdot |\Phi_i(p)|)),
\end{align}
where $\Phi_i(p)=s+\Delta s$ denotes the predicted SDF of $p$ for object $O_i$, $n$ denotes the normal of $p$, $\nabla$ denotes the spacial gradient of a 3D field, $\Omega$ denotes 3D space, and $\mathcal{S}_i$ denotes the shape surface of $O_i$. The first term is a regression loss for SDF. The second term is used to learn correct normals for points on the shape surface. The third term is derived from Eikonal equation to enforce the norm of the spatial gradient to be 1. The last term penalizes SDF values close to 0 for non-surface points, and $\delta$ is set to 100. The weights for four terms are set to $3e3$,$1e2$,$5e1$ and $5e2$ respectively. 

The normal consistency prior is defined as:
\begin{equation}
    L_{normal} = \sum_i \sum_{p \in \mathcal{S}_i}(1-\langle \nabla T(R_ip+v),n \rangle),
\end{equation}
where $\nabla T$ denotes the spatial gradient of template field $T$, and $R_i$ is the predicted rotation matrix for object $O_i$. This term enforces normals of points deformed from the same template point to be consistency.

The deformation smooth prior is defined as:
\begin{equation}
    L_{smooth} = \sum_i \sum_{p \in \Omega} \sum_{d \in \{X,Y,Z\}} \| \nabla D^v_i|_d(R_ip)  \|,
\end{equation}
where $\nabla D^v_i|_d$ denotes the gradient of the deformation field along axis $d$ for $O_i$. It penalizes the spatial gradient of the deformation field along $X$, $Y$ and $Z$ directions.

The minimal correction prior is defined as:
\begin{equation}
    L_c = \sum_i \sum_{p \in \Omega} | D^{\Delta s}_i(R_ip) |,
\end{equation}
where $\nabla D^{\Delta s}_i$ denotes the gradient of the correction field for $O_i$.

The shape code regularization loss is defined as:
\begin{equation}
    L_{reg} = \sum_i \|\bar \alpha_i\|^2_2.
\end{equation}

\paragraph{Training data.} We train our network with randomly rotated shapes from ShapeNet~\cite{shapenet2015} and OakInk dataset~\cite{yang2022oakink} as mentioned in the main paper. All meshes are converted to watertight manifold surfaces following the method described in~\cite{huang2018robust}. During training, we sample $2048$ points on each object surface with their normals as the input of network and the ground truth for normal consistency loss. We also sample 2048 spatial points uniformly as query points and compute their ground truth SDF values for SDF regression loss. 

\section{More results.}
\begin{figure}[!t]
        \centering
    \includegraphics[width=\columnwidth]{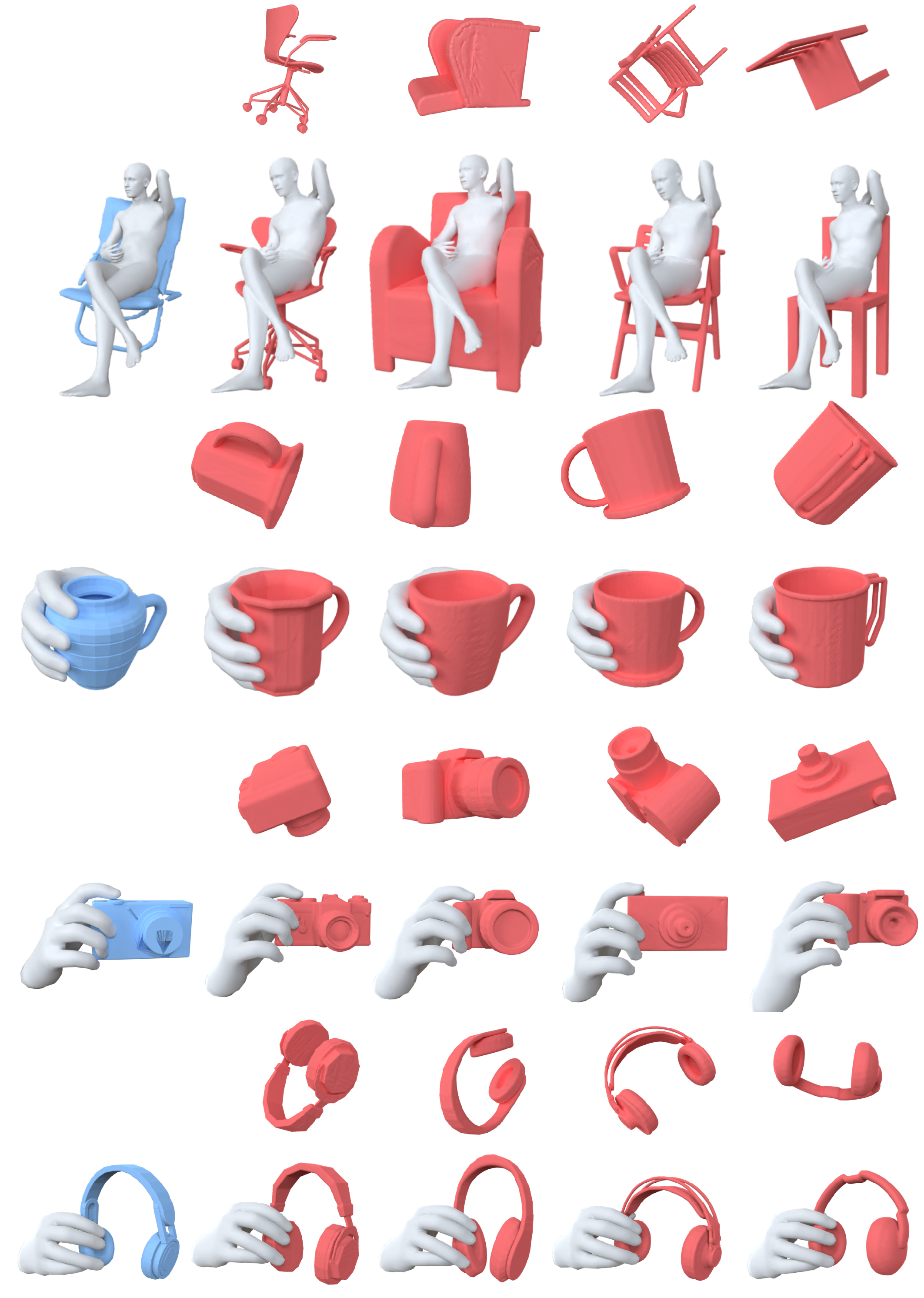}
    \caption{More interaction transfers on the random rotated objects.}
    \label{supp:static}
\end{figure}

\begin{figure*}[!t]
        \centering
    \includegraphics[width=0.9\linewidth]{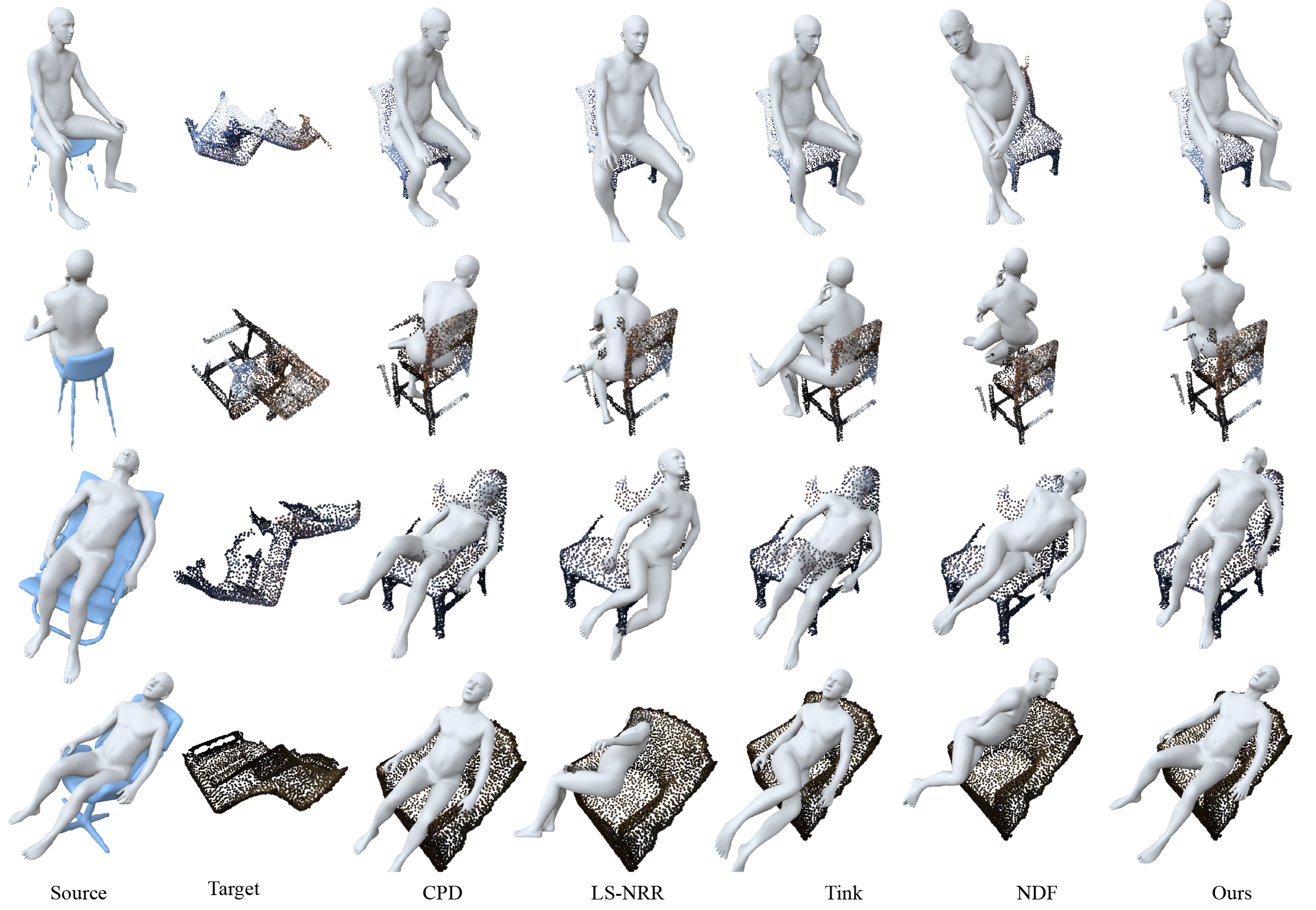}
    \caption{More comparisons on the real scan inputs}
    \label{supp:scan}
\end{figure*}

\begin{figure*}[!t]
    \centering
    \includegraphics[width= 0.9\textwidth]{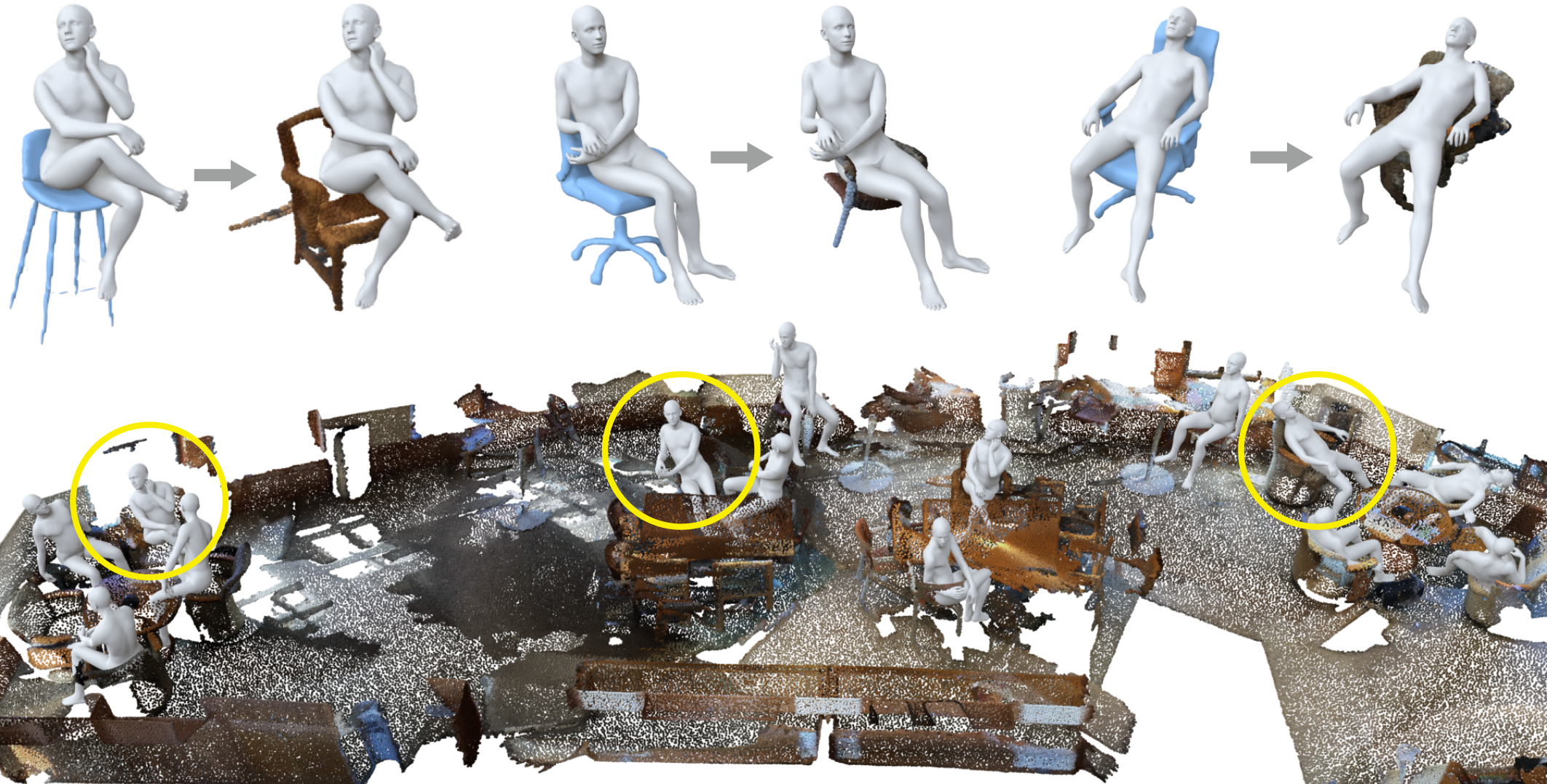}
    \caption{More examples of our method on real scan inputs. }
    \label{fig:more_scan}
\end{figure*}

\begin{figure*}[!t]
    \centering
    \includegraphics[width= \textwidth]{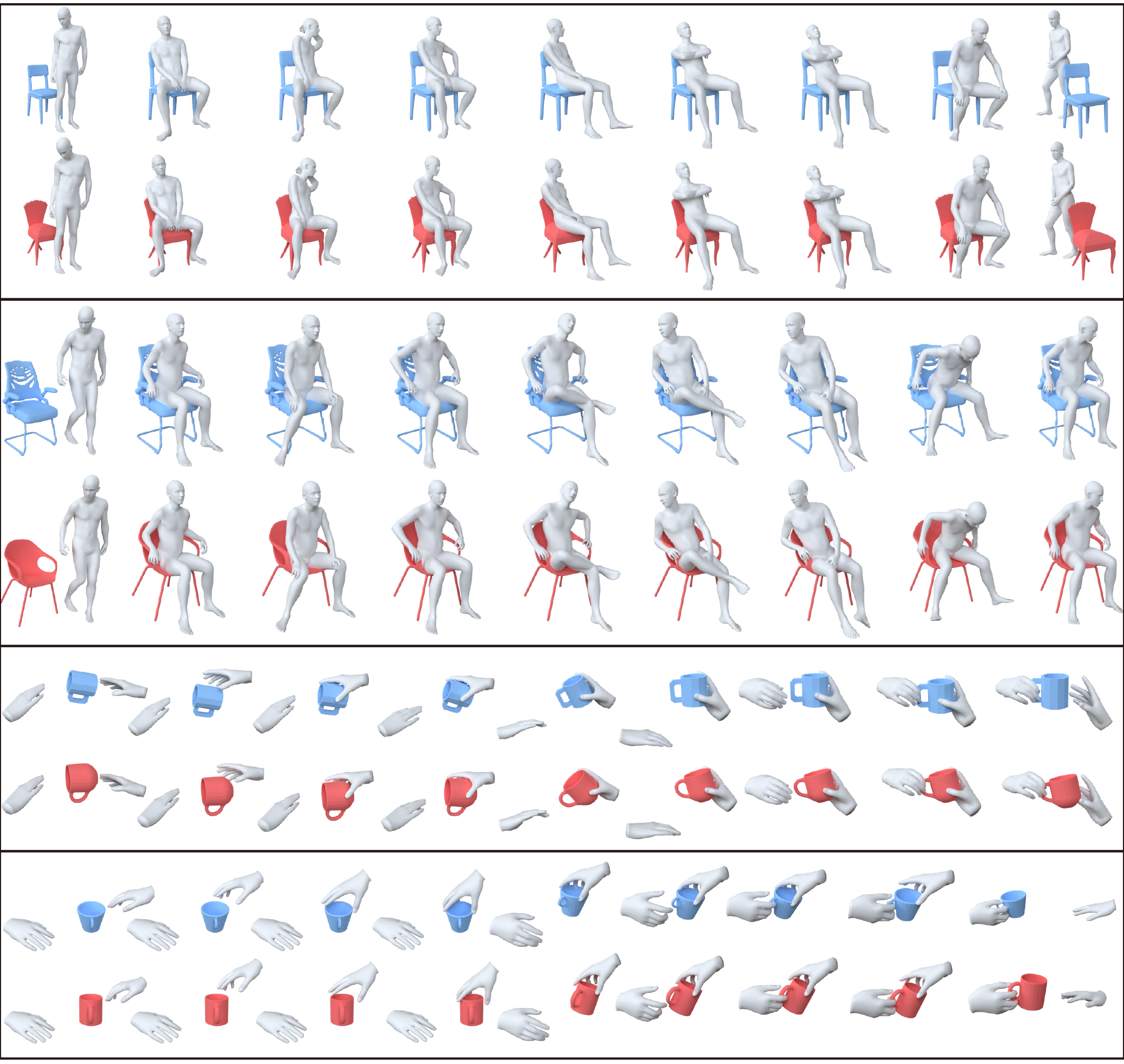}
    \caption{More results of our method on sequence transfer.}
    \label{fig:more_motion}
\end{figure*}

We provide more interaction transfer results on static interaction in Figure~\ref{supp:static}, on real scan inputs in Figure~\ref{supp:scan} and~\ref{fig:more_scan}, and on interaction sequences in Figure~\ref{fig:more_motion}. 

\clearpage
\end{document}